\documentclass{article} 
\usepackage{iclr2024_conference,times}

\usepackage{amsmath,amsfonts,bm}

\def\eqref#1{equation~\ref{#1}}

\def\1{\bm{1}}

\def\vz{{\bm{z}}}

\def\mA{{\bm{A}}}

\def\mC{{\bm{C}}}

\def\mK{{\bm{K}}}

\def\mQ{{\bm{Q}}}

\def\mV{{\bm{V}}}

\DeclareMathAlphabet{\mathsfit}{\encodingdefault}{\sfdefault}{m}{sl}
\SetMathAlphabet{\mathsfit}{bold}{\encodingdefault}{\sfdefault}{bx}{n}

\def\gC{{\mathcal{C}}}

\DeclareMathOperator*{\argmin}{arg\,min}

\usepackage[colorlinks, linkcolor=black, anchorcolor=black, citecolor=blue]{hyperref}
\usepackage{url}
\usepackage{graphicx}
\usepackage{subcaption}
\usepackage{bm}
\usepackage{multirow}
\usepackage{array}
\usepackage{enumitem}
\usepackage{tabularx}
\usepackage{booktabs} 
\usepackage[linesnumbered,ruled,vlined]{algorithm2e}
\SetKw{KwBy}{by}
\usepackage{wrapfig}
\usepackage{caption}
\usepackage{float} 
\usepackage{makecell}
\newcommand{\tabincell}[2]{\begin{tabular}{@{}#1@{}}#2\end{tabular}} 
\title{Model Tells You What to Discard: \\Adaptive KV Cache Compression for LLMs}

\author{Suyu Ge$^{1}$\thanks{Authors contributed equally to this research. Code is available is at https://github.com/machilusZ/FastGen}, Yunan Zhang$^{2*}$, Liyuan Liu$^{2*}$, Minjia Zhang$^{2}$, Jiawei Han$^{1}$, Jianfeng Gao$^{2}$\\
$^{1}$University of Illinois Urbana-Champaign, $^{2}$Microsoft\\
\texttt{\{suyuge2,hanj\}@illinois.edu}\\
\texttt{\{yunanzhang,lucliu,minjiaz,jfgao\}@microsoft.com}\\
}
\newcommand{\ours}{FastGen}

\iclrfinalcopy 
\begin{document}

\maketitle

\begin{abstract}
In this study, we introduce adaptive KV cache compression, a plug-and-play method that reduces the memory footprint of generative inference for Large Language Models (LLMs).
Different from the conventional KV cache that retains key and value vectors for all context tokens, we conduct targeted profiling to discern the intrinsic structure of attention modules.
Based on the recognized structure, we propose FastGen, which constructs the KV cache in an adaptive manner: evicting long-range contexts on attention heads emphasizing local contexts, discarding non-special tokens on attention heads centered on special tokens, and only employing the standard KV cache for attention heads that broadly attend to all tokens.
Moreover, with the lightweight attention profiling used to guide the construction of the adaptive KV cache, {\ours} can be deployed without resource-intensive fine-tuning or re-training.
In our experiments across various asks, {\ours} demonstrates substantial reduction on GPU memory consumption with negligible generation quality loss. 

\end{abstract}


\section{Introduction}

Based on the Transformer architecture, autoregressive language models have attracted extensive attention~\citep{openai2023gpt4,touvron2023llama}. 
Along with the increase of model size, these models present significant challenges in terms of computational complexity and GPU memory consumption~\citep{540b}.
Since these models achieve remarkable success across diverse applications, there is a pressing need for serving these models in an economically feasible manner.

The generative inference of LLMs usually involves using the \emph{KV Cache} mechanism
to improve the generation speed. KV cache stores previously computed Key/Value vectors in attention calculation and reuses those values for the current token generation. As such, it avoids recalculations of previous tokens at each token generation step at the cost of extra memory consumption. Despite being a prominent technique, the memory consumption of KV cache increases rapidly as the model size and generation length increase, drastically increasing the pressure of on-device memory.  

When memory usage exceeds GPU capacity, the generative inference of LLMs typically resort to offloading~\citep{Aminabadi2022DeepSpeedIE,Sheng2023HighthroughputGI}.
While these methods help mitigate the pressure on the scarce GPU memory from using KV cache, offloading KV cache to CPU/NVMe can still add non-trivial overhead to generative inference performance due to the limited PCIe bandwidth between the GPU and CPU on many devices. 
Therefore, it becomes a crucial task to reduce the memory footprint of KV cache without costly retraining or fine-tuning. 



\begin{figure}[h]
\includegraphics[width=0.9\textwidth]{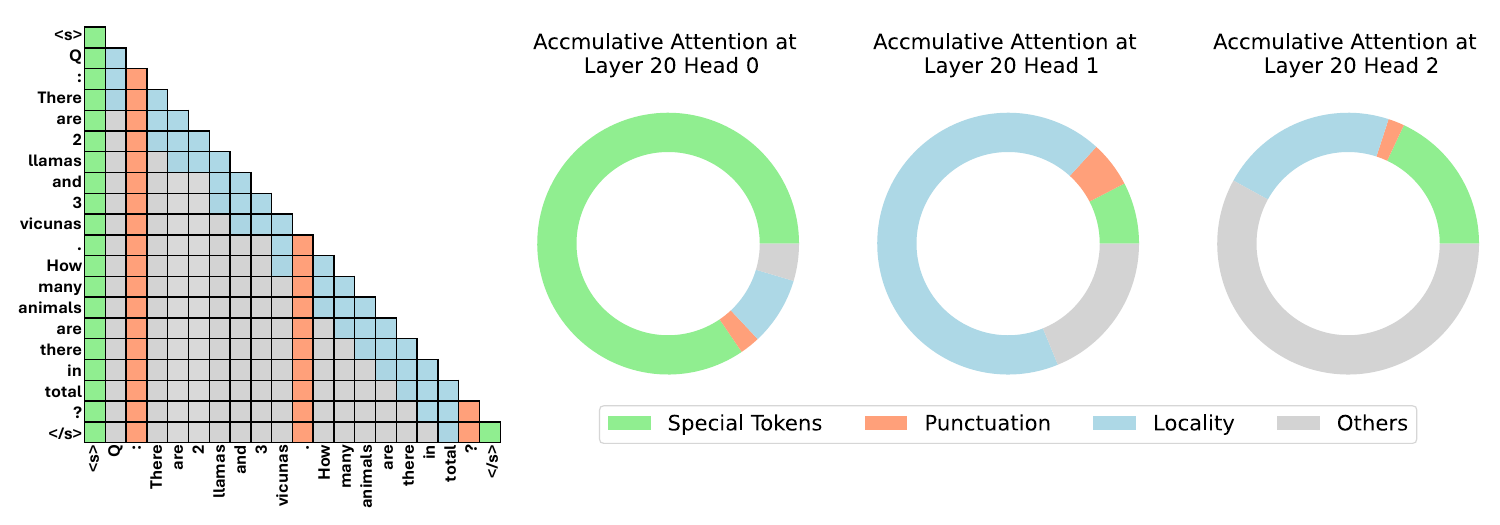}
\caption{Different attention heads usually have different structures. Left: Four common attention structures 
 (more details are elaborated in Section~\ref{sec:method} and Section~\ref{sec:analyses}). Right: Attention map compositions of three attention heads that are in the same layer.}
\label{fig:obs}
\end{figure}

Our study starts from the observation (Figure \ref{fig:obs}) that 
there are abundant structures observed in attention modules \citep{sixteen, lift-prune,bert-look,structureprune,sparsetransformer}, and not all attention modules need to attend to all tokens \citep{deja,h2o,scissorhands}.
Intuitively, harvesting such structures and compressing cached vectors could substantially reduce memory consumption and accelerate text generation.

Based on this intuition, we propose {\ours} to \emph{accelerate the generative inference by adaptively compressing the KV cache on the fly}. 
First, we employ an efficient profiling algorithm to recognize the structural patterns for attention modules. 
Under the guidance of this profiling, we then construct the KV cache for various modules adaptively. 
With this diagnose-before-compress approach, {\ours} effectively reduces the memory footprint of KV cache while preserving the model quality. 

\begin{minipage}[t]{0.49\textwidth}

\begin{algorithm}[H]
\DontPrintSemicolon
\KwIn{Feasible Policy Set ($\gC$), Prompt}
\KwOut{Adaptive KV Cache}
\For{Attention Head $H_i$ in LLM} {
    $\mK^i, \mQ^i, \mV^i \gets H_i(\mbox{Prompt})$ \;
    $\mA^i \gets \mbox{softmax}({\mQ^i \mK^i}^T )$\;
    $\mC^i \gets $ apply Equation~\ref{eqn:opt} to $\mA^i$ \;
    \tcc{C$^i$: optimal policy}
    $\mK^i_{\mC^i}, \mV^i_{\mC^i} \gets f(\mK^i, \mV^i, \mC^i)$\;
    $\hat{\mK}^i, \hat{\mV}^i \gets \mK^i_{\mC^i} \mV^i_{\mC^i}$ \;
}
\Return{$\{\mC^i, \hat{\mK}^i, \hat{\mV}^i\}$}
\caption{FastGen--Prompt Encoding. }
\label{algo:encoding}
\end{algorithm}

\end{minipage}
\hfill
\begin{minipage}[t]{0.49\textwidth}

\begin{algorithm}[H]
\DontPrintSemicolon
\KwIn{Adaptive KV cache ($\{\mC^i, \hat{\mK}^i, \hat{\mV}^i\}$)}
\KwOut{Generated Text}
$\vz_0 \gets$ last prompt token \; 
\For{$j\in \{1, \cdots, \mbox{Max Generate Length} \}$ }{
    \For{Attention Head $H_i$ in LLM} {
        $\mK^i, \mQ^i, \mV^i \gets H_i(\vz_{j-1}, \hat{\mK}^i, \hat{\mV}^i)$ \;
        $\mK^i_{\mC^i}, \mV^i_{\mC^i} \gets f(\mK^i, \mV^i, \mC^i)$\;
        $\hat{\mK}^i, \hat{\mV}^i \gets \mK^i_{\mC_i}, \mV^i_{\mC_i}$ \;
    }
    $\vz_j \gets $ sample from LLM prediction \;
}
\Return{$\{\vz_j\}$}
\caption{FastGen--Token Generation. }
\label{algo:decoding}
\end{algorithm}

\end{minipage}

\begin{figure}[h]
\includegraphics[width=\textwidth]{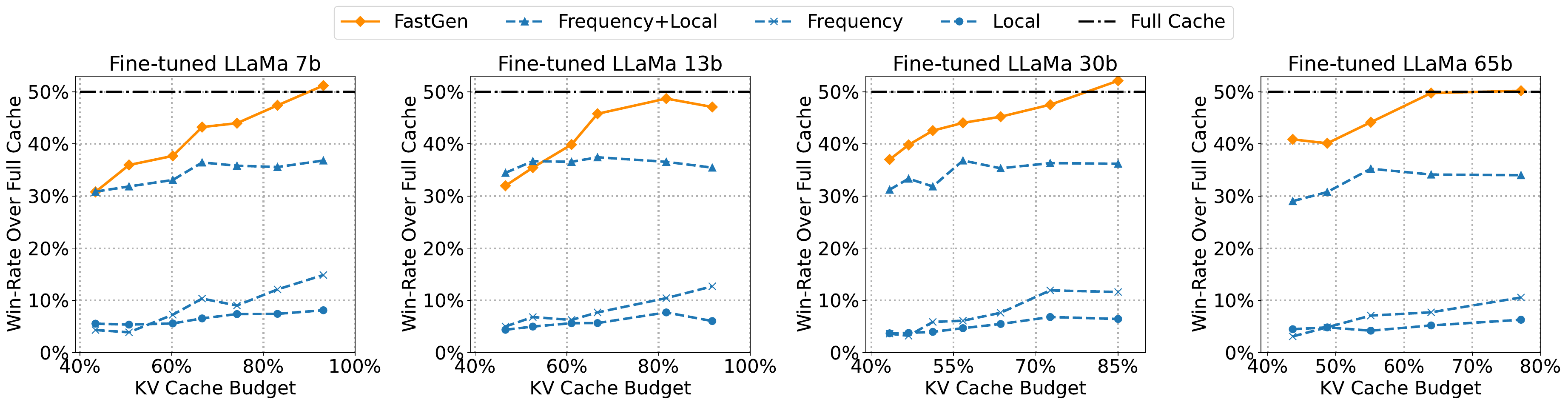}
\caption{Performance of Adaptive KV Cache (FastGen) and Fixed KV Cache (Frequency, Local, and Frequency+Local; \citeauthor{h2o}, \citeyear{h2o} and \citeauthor{scissorhands}, \citeyear{scissorhands}) on AlpacaEval.}
\label{fig:exp_main}
\end{figure}

In our study, {\ours} recognizes five fundamental attention structures and applies them correspondingly. 
Specifically, some attention modules mostly attend to local contexts, for which we construct a KV cache that evicts long-range contexts; some primarily attend to specific tokens/punctuations, for which we create a KV cache that retains only special tokens/punctuations; some have attention maps that are column-wise sparse, for which we discard the least frequently attended tokens; and some broadly attend to all tokens, for which we employ the standard KV cache and store all tokens.

In this way, {\ours} is able to compress the KV cache while retaining the original functionality of attention modules.
Remarkably, 
{\ours} does not require any fine-tuning and can be 
applied in a plug-and-play manner. 
This is a big advantage of {\ours}, because the training cost on extra-large models~\citep{gpt-3}, can hardly be afforded by many research labs or practitioners. 

We evaluate {\ours}  on Llama 1 \citep{touvron2023llama} with a suite of major benchmarks covering generative tasks in math, code, knowledge, and common sense reasoning. 
{\ours} effectively performs KV cache compression with negligible generation quality loss (i.e., recover over 95\% of attention scores with 35\% cache compressed).
Notably, as to the 30b model in Figure~\ref{fig:exp_main}, {\ours} (50\% cache compressed) surpasses all fixed KV compression methods (15\% cache compressed).

    


\section{Related Work}
\label{sec:related}

\textbf{Token Dropping and KV Cache Compression.}
Many efforts have been made to improve the model efficiency for LLMs. 
For recurrent neural networks, one method is to skip multiple tokens at a given time step \citep{Campos2017SkipRL,Seo2017NeuralSR,Hansen2019NeuralSR}.
Since Transformer models quickly attracted lots of attention, 
\citet{powerbert} proposes to eliminate redundant words in BERT \citep{bert} based on their attention scores, while \citet{funnel} compresses the input sequence by adding pooling layers to the encoding modules of the transformer architecture. 
Recently, \citet{pyramid} adds a token selection task to the original BERT model that learns to select performance-crucial tokens, and \citet{tokenprune} designs a learnable threshold to detect unimportant tokens to prune. 
Meanwhile, many efforts have been made to explore the possibility of compressing the hidden state of tokens rather than explicitly reducing the sequence length~\citep{transkimmer, hashexit, patience}.

Nevertheless, these methods can only be applied to non-autoregressive models and typically require an additional re-training phrase, making them less suitable for auto-regressive LLMs like ChatGPT and Llama.
Recognizing this gap, researchers started examining the potential of pruning tokens within the KV cache of auto-regressive LLMs. 
\citet{gist} learns to compress the prompts into a few special tokens to reduce memory pressure during caching.
However, the token prediction requires model re-training and could be an expensive overhead during inference.
Meanwhile, several concurrent methods propose to leverage accumulated attention score as the criteria to identify important tokens in the KV cache \citep[e.g.,][]{Sheng2023HighthroughputGI,h2o,scissorhands}.
Instead of investigating a specific eviction policy, this study aims to synergistically coordinate diverse eviction policies to better align with model-specific attributes.

\textbf{Underlying Structure of Attention.} 
Inspired by the success of Transformer, extensive studies have been conducted to explore the underlying mechanism of different self-attention heads.
\citet{lift-prune} analyzed the self-attention heads in BERT using LRF \citep{lrf} and characterized them into interpretable roles, one of which is attending adjacent tokens all the time.
\citet{sixteen} demonstrated that heads in the same layer could have different impact on the performance while the importance of each head changes across tasks.
\citet{bert-look} and \citet{dark} identified such patterns as some heads primarily attend to separator tokens, adjacent tokens and a combination of these. 
While most previous studies mainly considered encoder models, FastGen is motivated by consistent patterns we have observed in decoder-only models. 
Like previous studies, FastGen also explores the structure of the attention mechanism to improve inference efficiency, but focusing on characterizing the KV cache of different attention heads.

\section{Adaptive KV Cache Compression}
\label{sec:method}

In this section we first introduce the problem formulation, and then present attention profiling and adaptive KV cache compression. 

\subsection{Generative Inference of Autoregressive LLMs}

A typical generative model inference involves two steps: prompt encoding and token generation. 

\textbf{Prompt Encoding. } 
When an autoregressive transformer-based LLM generates the $i$-th token, the attention module needs to refer to all the preceding $i-1$ tokens, i.e., the key and value vectors (KV vectors) of these tokens. 
To circumvent redundant KV vector computations when generating succeeding tokens, all KV vectors are stored in the \emph{KV cache} once they are generated. 

\textbf{Token Generation. }
Once prompt encoding is finished, the LLM generates the output token by token. 
At each generation step, the LLM needs to encode the new token(s) generated in the previous step. 
After a new token is generated, its associated KV vectors are appended to the current KV cache.
Thus, the size of KV cache increases linearly with the number of tokens being generated.

\subsection{{\ours}} 
As described in Section~\ref{sec:related}, many previous studies of compressing KV cache for improving inference efficiency do not leverage the intricate attention structure in LLMs. As to be detailed in Section~\ref{sec:analyses}, attention heads in LLMs often function distinctively, indicating the need for tailoring the compression strategy to each individual attention head.

With these insights, we introduce FastGen: a dual-phase algorithm for crafting an adaptive KV cache. 
During the prompt encoding phase, model profiling is conducted to discern the behavior of various attention heads, so that we can choose the most appropriate compression strategy for each head.
Then, in the token generation phase, instead of indiscriminately appending new KV vectors for each newly generated token, we manage the KV cache based on the selected compression strategies. 



\subsection{Model Profiling}
\label{subsec: profiling}

Model profiling is conducted based on the result of prompt encoding. 
Specifically, for a compression policy $\mC$, we mark the corresponding KV cache compression as $\mK_\mC, \mV_\mC = f(\mK, \mV, \mC)$, where $\mK_\mC$ and $\mV_\mC$ are the compressed KV cache. 
Then, for attention map $\mA=\mbox{softmax}(\mQ \mK^T)$, we pick the optimal policy that can recover $\mA$ with a recover ratio $T$ with the minimum memory cost:

\begin{equation}po
\label{eqn:opt}
\mC^* = \argmin_{\mC \in \gC}\; \mbox{CacheMemoryCost}(\mC) \;\; \mbox{ s.t. } \;\; |\mA - \mbox{softmax}(\mQ \mK_{\mC}^T)| \leq 1 - T,
\end{equation}
where $\gC$ is the set of all feasible compression policies, $\mbox{CacheMemoryCost}(\mC)$ is the target KV cache budget of the compression policy $\mC$, and $T$ is a predefined hyper-parameter representing how much we want the policy to recover $\mA$.
As to be discussed in Section~\ref{sec:exp}, {\ours} is able to recover +95\% of the attention map with +40\% compression ratio for a 65B model.
The final prompt encoding algorithm that includes model profiling is presented in Algorithm~\ref{algo:encoding}.

Intrinsically, our method assumes that the structure of the attention map for a head is stable through the generation process. So, it is sufficient to use only the encoded prompt to select a proper compression policy.
It is worth noting that existing literature has provided theoretical justifications for using solely encoded prompts to capture attention structures for the full contexts~\citep{h2o,scissorhands}. 
In our study, we also empirically verified this, as to be elaborated in Section~\ref{sec:analyses}.

\subsection{KV Cache Compression Policies}
\label{subsec:compression_policy}
In our experiments we observe that a large number of attention heads closely follow certain patterns, as to be detailed in Section~\ref{sec:analyses}. 
Thus, in addition to the conventional full KV cache policy, we also consider four fundamental KV cache compression policies. 
While we mainly use these four fundamental KV cache compression policies for evaluation in this study, it is easy for FastGen to use numerous other strategies. 
The four KV cache compression policies are:
\begin{itemize}[leftmargin=*]
    \item
    \textbf{Special Tokens.} We keep in KV cache only special tokens, such as the begin-of-the-sentence token $<$s$>$, the instruction token $[$INST$]$, and so on. This policy is referred to as $\mC_{\small \mbox{special}}$. 
    
    \item 
    \textbf{Punctuation.} We keep in the KV cache only punctuation tokens like ".", ":", "?". This policy is referred to as $\mC_{\small \mbox{punct.}}$. 

    \item 
    \textbf{Locality}
    This policy evicts long-range contexts. Once the relative distance between the context token and the current token exceeds a threshold, the KV cache of the context token will be evicted. The threshold is determined by a pre-defined ratio $r_l$ of the length budget of local context over the input sequence length. This policy is referred to as $\mC_{\small \mbox{local}}$. 

    \item 
    \textbf{Frequency (Heavy Hitter)} This policy has been used in multiple previous studies~\citep[e.g.,][]{Sheng2023HighthroughputGI,h2o,scissorhands}. We monitor for each token its cumulative sum of attention score, then treat these scores as token \emph{frequency} and only keep the most frequent tokens in the KV cache.  
    The length budget of frequent tokens over the current sequence length is controlled by a ratio $r_f$. This policy is referred to $\mC_{\small \mbox{frequent}}$.
\end{itemize}

\paragraph{Hybrid Policies.} 
In practice, it is often necessary to use hybrid policies that combines the aforementioned compression policies.
Since the total number of hybrid policies is hugh,
in our study we use a greedy method to construct a small set of hybrid-policies as follows
\begin{eqnarray}
\label{eqn:policyset}
    \gC = \{\mC_{\small \mbox{special}}, \mC_{\small \mbox{special+punct.}}, \mC_{\small \mbox{special+punct.+frequent}}, \mC_{\small \mbox{special+punct.+frequent+local}}, \mC_{\small \mbox{full}} \},
\end{eqnarray}
where the sum of two compression strategies is to compute the union of their compressed KV cache, and $\mC_{\small \mbox{full}}$ refers to full KV cache without compression.

We use $\mC_{\small \mbox{special}}$ as a component in all hybrid policies for two reasons: 1) We observe that high attention scores are usually allocated towards $\mC_{\small \mbox{special}}$, as to be detailed in Section~\ref{sec:analyses}, indicating that $\mC_{\small \mbox{special}}$ are crucial for attention map recovery; 2) the compressed cache of $\mC_{\small \mbox{special}}$ is memory-efficient since there are usually less than 5 special tokens in a sentence.
In other words, it brings little-to-no extra memory cost by always including $\mC_{\small \mbox{special}}$.
Similarly, $\mC_{\small \mbox{punct.}}$ is often used as a component to form hybrid policies due to its memory-efficiency, i.e., the number of punctuations in a sentence is small.
The final algorithm for token generation is presented in Algorithm~\ref{algo:decoding}.

\section{Diversity and Stability of Attention Structures}
\label{sec:analyses}

In this section we present an empirical study to show the effectiveness of adaptive KV cache compression. 
First, we demonstrate that different attention heads typically possess distinct structures.
Then, we show that these attention head structures remain relatively consistent across different attention heads at different positions.
We do so by analyzing the attention scores of Llama 1 65B using random samples from GSM8k~\citep{gsm8k}. 


\subsection{Head Distinctive Attention Structure}
\begin{figure}[H]
\centering
\includegraphics[width=\textwidth]{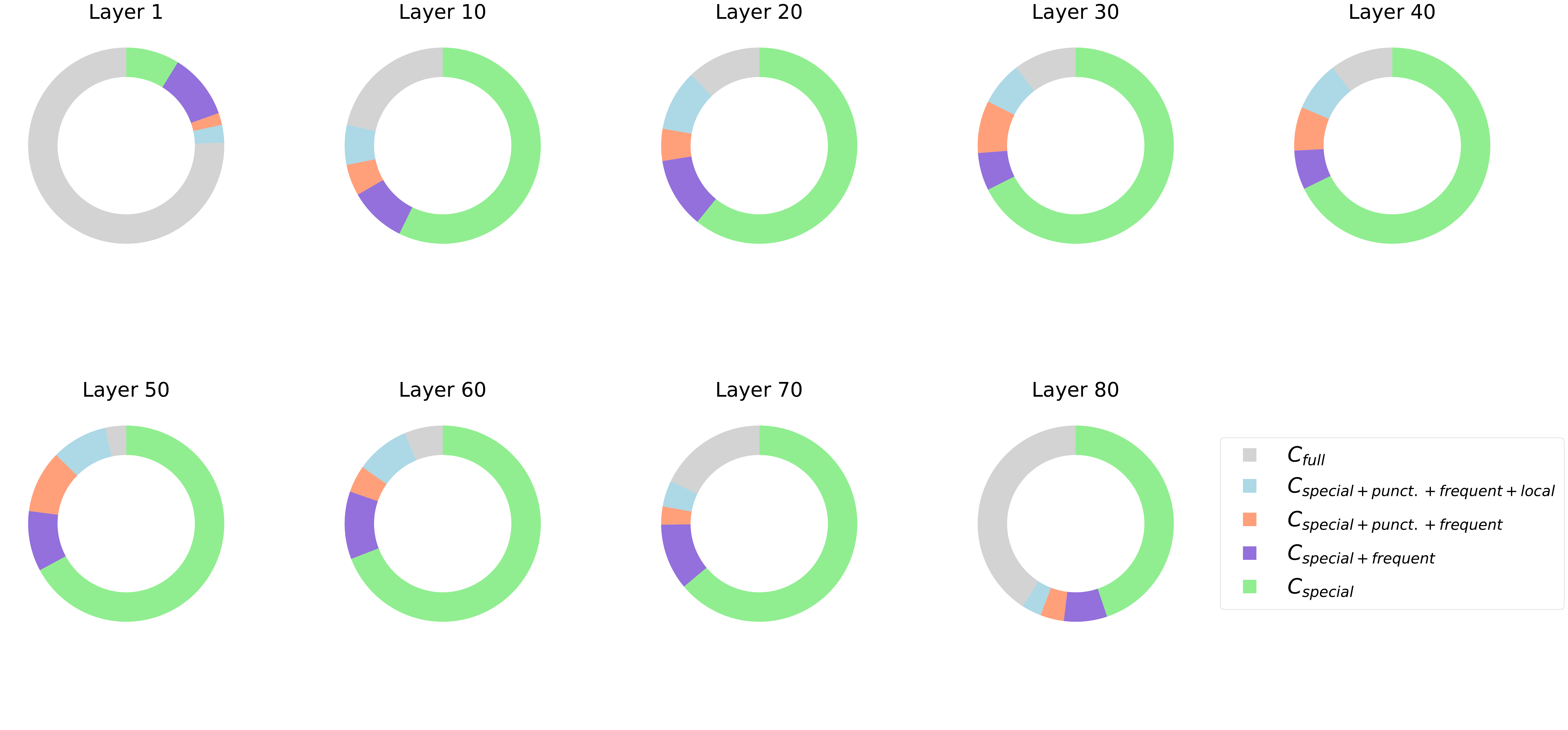}
\vspace{-1cm}
\caption{Attention profiling result distribution across different layers.}
\label{fig:diverse}
\end{figure}
\vspace{-0.2cm}

\paragraph{Setting.} 
We perform model profiling with a recover threshold of $0.95$ and compute the distribution of profiling results for $\{1, 10, 20, 30, 40, 50, 60, 70, 80\}$ layers. 
The result is shown in Figure~\ref{fig:diverse}. 

\paragraph{Observation.} 
Figure~\ref{fig:diverse} shows that attention heads in different layers have vastly different structures.  
Specifically, for the initial and final layers, they have more attention heads assigned to the full KV cache, indicating attention heads in these layers are likely to attend to all tokens. 
Meanwhile, for middle layers, the attention map focuses on special tokens, indicating that most attention heads of these layers primarily attend to special tokens (i.e., the accumulated attention score on special tokens is higher than $0.95$ for these attention heads). 
Figure~\ref{fig:obs} shows the structure of different attention heads in the same layer. We see that attention structures differ across different layers and heads. 

These results indicate that it is suboptimal to apply the same KV cache to all layers without adaptation, and that it is beneficial to detect the structure of each attention head so as to select the optimal compression policy to construct the KV cache. 

\subsection{Profile Tends to Be Consistent in One Sequence}
The previous section demonstrates the great potential for constructing adaptive KV cache in accordance with the structure of different attention heads. 
Here, we show for each instance, it is sufficient to  conduct one-shot model profiling, as outlined in  Section~\ref{subsec: profiling}. 
Specifically, for a given prompt, we show that the attention structure of each head remains consistent through the decoding process. 

\begin{figure}[H]
\centering
\includegraphics[width=\textwidth]{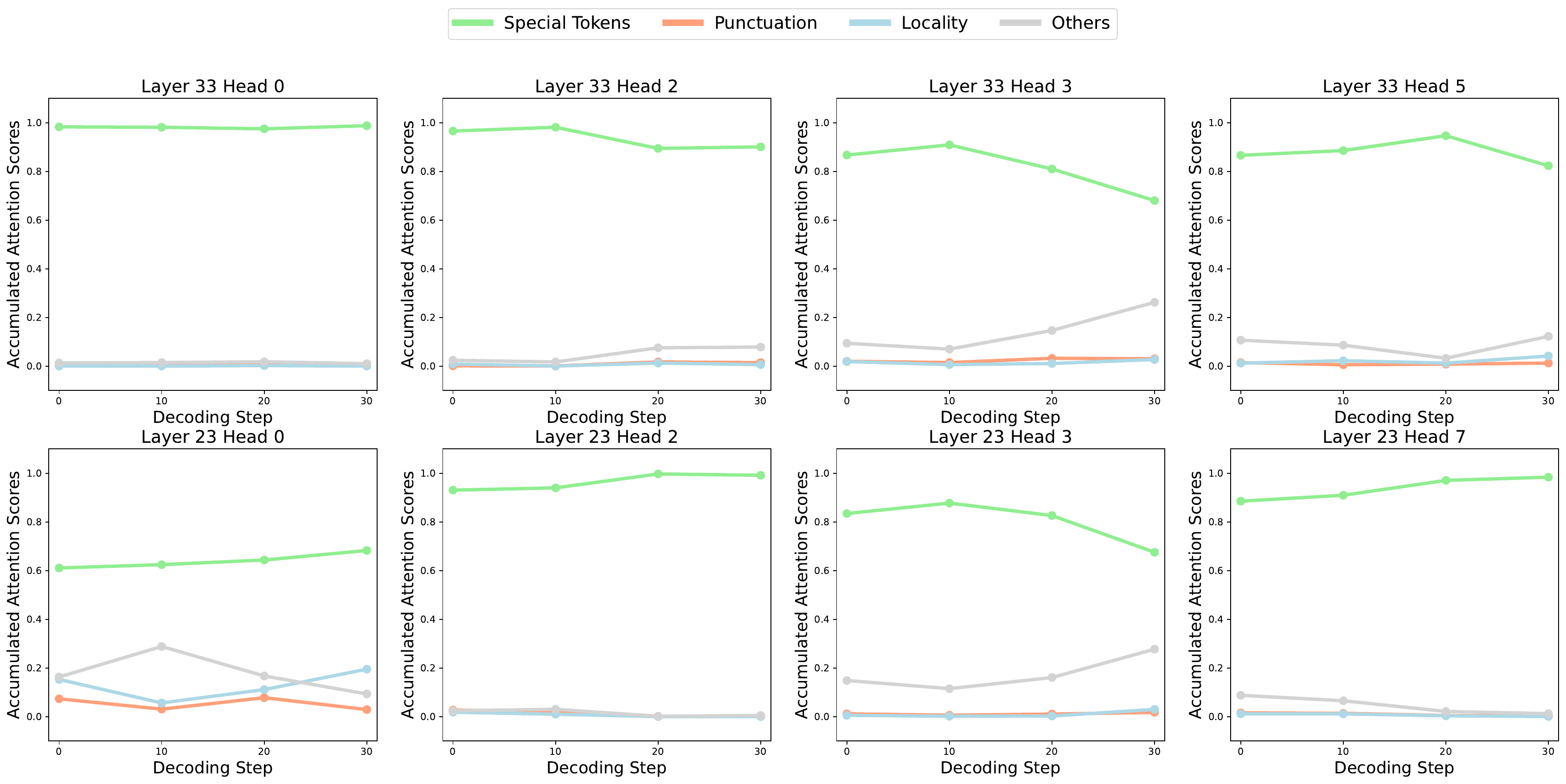}
\caption{Accumulated attention score at 1st (prompt encoding), 10th,  20th, 30th decoding steps.}
\label{fig:heads}
\end{figure}
\vspace{-0.3cm}

\paragraph{Setting.} 
Following Figure~\ref{fig:obs}, we compute the accumulated attention score for attention heads in different layers of Llama 1 65B at multiple decoding steps (i.e., 1st, 10th, 20th, 30th). 
We visualized the resulting accumulated score in Figure~\ref{fig:heads}.

\paragraph{Observation.} 
Despite some fluctuations of accumulated attention scores across time steps, 
the pattern of the attention maps remains relatively stable. 
For example, Layer 33 Head 0 and Layer 23 Head 2 almost only attend to the special token, while the locality and punctuation plays an important role in Layer 23 Head 0. 
As to Layer 23 Head 3, more than 10\% of the attention score is allocated to the others portion, making it suitable for a uncompressed KV cache $\mC_{\small \mbox{full}}$. 

In addition, we observe that a large portion of attention scores are on special tokens in all cases. This justifies the greed method we used to construct hybrid policies, as described in Section~\ref{subsec:compression_policy}.



\section{Experiment}
\label{sec:exp}
We conduct comprehensive experiments to demonstrate the effectiveness of {\ours} on memory footprint reduction and generation quality preserving.
First, we report the trade-off between memory reduction and end-to-end generation quality in Section~\ref{subsec:endtoend}, and discuss the compression ratio of {\ours} in Section~\ref{subsec:memory}.  
To demonstrate the superiority of FastGen on real-world systems, we demonstrate the end-to-end latency change in Section~\ref{sec:latency} and the profiling overhead in Section~\ref{sec:profiling_cost}.
Finally, we present ablation studies and discussions in Section~\ref{subsec:ablation}.

\subsection{Trade-off between performance and memory reduction}
\label{subsec:endtoend}

\begin{figure}[h]
    \includegraphics[width=\textwidth]{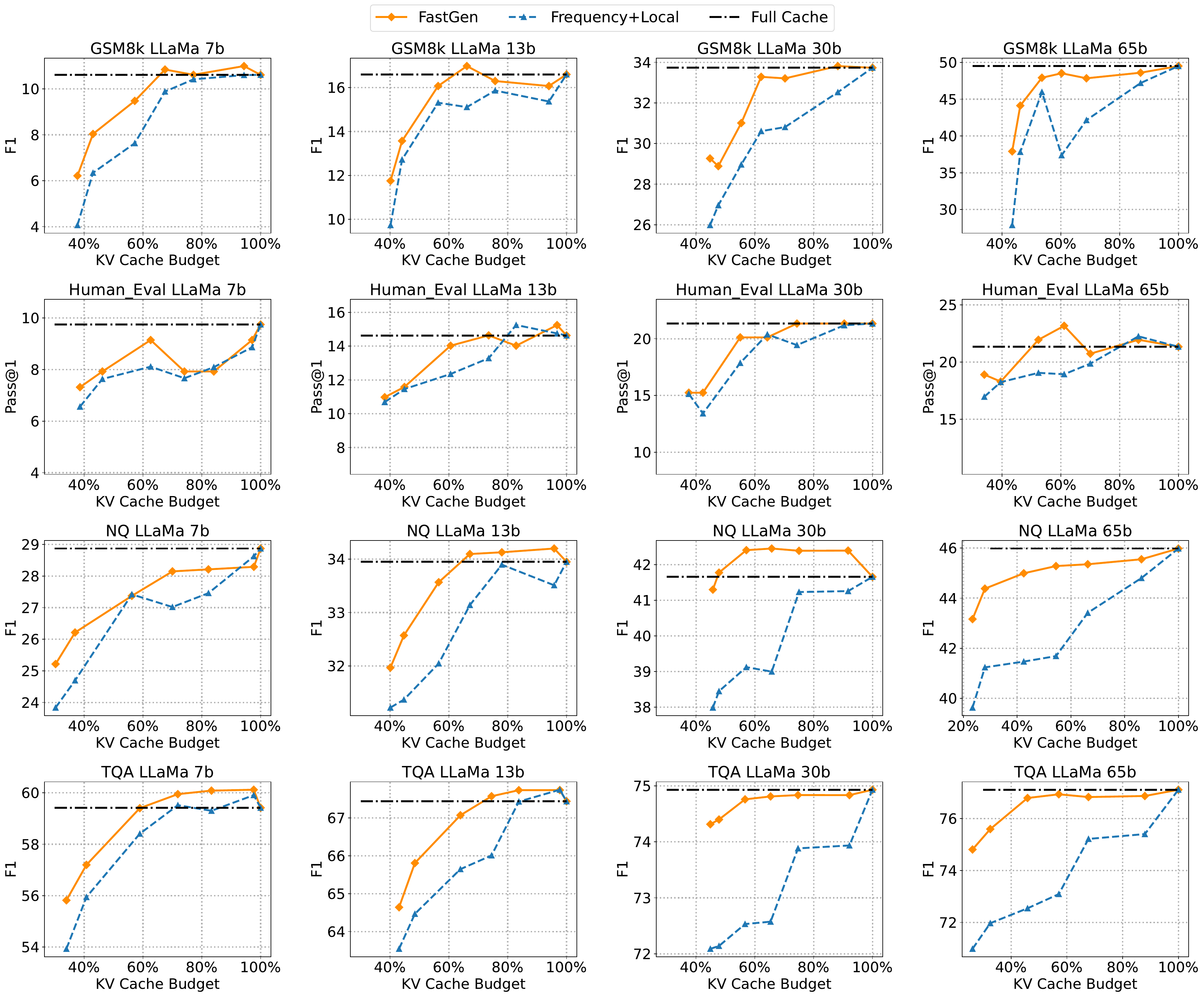}
    \caption{Performance of Adaptive KV Cache (FastGen) and Fixed KV Cache (Frequency+Local; \citeauthor{h2o}, \citeyear{h2o} and \citeauthor{scissorhands}, \citeyear{scissorhands}) of Llama 1 on GSM8k, HumanEval, NQ, and TQA.}
    \label{fig:exp_main_base}
\end{figure}
\vspace{-0.3cm}

\paragraph{Backbones.} 
We conduct experiments with both Llama 1 \citep{touvron2023llama1} and its fine-tuned variants, with model sizes ranging from 7B to 65B.
For fined-tuned variants, we do not choose the open-sourced Llama 2-chat \citep{touvron2023llama} model due to its grouped-query attention techniques.
Instead, we use the original multi-head attention architecture in this study and leave the integration of grouped-query attention to future work.
To prepare a comparable instruction-following model for analysis, we fine-tuned the Llama 1 model with open-sourced instruction-tuning datasets.
Specifically, the fine-tuned variants are trained on LIMA\footnote{https://huggingface.co/datasets/GAIR/lima.} data \citep{lima} and Open Assistant\footnote{https://huggingface.co/datasets/OpenAssistant/oasst1.} \citep{openass} data.

\paragraph{Tasks.}
We use standard generation tasks to evaluate Llama 1 and our fine-tuned Llama 1 models.
For Llama 1, we choose 4 different tasks, including HumanEval~\citep{human_eval}, GSM8k~\citep{gsm8k}, NQ~\citep{nq} and TQA~\citep{tqa} to evaluate models' abilities on different domains (code, math, question answering and reading comprehension).
Note that in the four tasks, each testing sample is in a generative format, where answers are extracted after model generation finishes. 
This is crucial for a fair comparison on model's generation quality.
We evaluate the instruction finetuned LLaMa model on the instruction tuning benchmark AlpacaEval \citep{alpaca_eval}, which consists of 805 question prompts from diverse domains. 

\paragraph{Experiment Setup.} 
The evaluation of the Llama 1 model follows the default setting and evaluation metrics on each benchmark.
We calculate F1 scores for GSM8k, NQ and TQA, and use the code execution Pass@1 rate for HumanEval.
While evaluating an instruction-tuning model remains challenging, we follow previous work~\citep{lima,touvron2023llama} to use GPT4 as an evaluator for pair-wise comparison between two different model generations.
For each prompt, we input the {\ours} generation and the generation from the same model with Full KV Cache as a pair, and ask GPT4 to judge which one is better.
We then calculate the win rate of {\ours} over Full Cache.
Hypothetically, the win rate of a lossless method should be around 50\%. 
Aside from full-cache models, we also include non-adaptive KV cache methods for comparison. 
Specifically, we apply $\mC_{\small \mbox{local}}$, $\mC_{\small \mbox{frequent}}$, and $\mC_{\small \mbox{local+frequent}}$ to all attention head without any adaptation, as baselines. 
It is worth mentioning that $\mC_{\small \mbox{local+frequent}}$ is a very strong baseline as it is identical to the H$_2$O method \citep{h2o} and the Scissorhands method \citep{scissorhands}. 
We set $r_l=0.3$, $r_f=0.3$ in {\ours}, and only change the recovery ratio $T$ to control the pruned KV cache ratio.
For generation, we use nucleus sampling~\citep{holtzman2019curious} with temperature T = 0.6, p = 0.9.
Experiments are conducted on 8 NVIDIA A100 80GB GPUs.

\paragraph{Main Results.} 
In Figure~\ref{fig:exp_main} and Figure~\ref{fig:exp_main_base}, we present the model quality as a function of KV cache budget increasing from $30\%$ to $100\%$.
For 30B models, {\ours} (50\% cache compressed) surpasses all non-adaptive KV compression methods (15\% cache compressed) .
Also, we can see {\ours} achieves more KV cache reduction ratio as the model size increases, while preserving the same model quality. 
For example, achieving a $45\%$ win rate, FastGen can get as much as  44.9\% pruned ratio on Llama 1-65B, compared to 16.9\% pruned ratio on Llama 1-7B.
In all settings, FastGen shows consistent and significant improvement over non-adaptive compression methods. 
The results validate the effectiveness of adaptive KV cache compression using {\ours}, despite its simplicity.


\begin{table*}[h]
\small
\centering
\resizebox{0.65\columnwidth}{!}{
\begin{tabular}{l | c ccc | c}
\toprule
\multirow{2}{*}{Model Size} & \multicolumn{4}{c|}{KV Cache} & \multirow{2}{*}{Win rate}\\
& Full & {\ours} & Pruned ratio & $T$ & \\ 

\midrule
\multirow{3}{*}{7B}
& \multirow{3}{*}{4.3Gb} & 1.9Gb & 56.6\% & 91\%  & 30.8\% \\ 
&                     & 2.6Gb & 39.8\% & 95\%  & 37.7\% \\ 
&                     & 3.6Gb & 16.9\% & 98\%  & 47.4\% \\ 

\midrule
\multirow{3}{*}{13B} 
& \multirow{3}{*}{6.7Gb}  & 3.1Gb  & 53.4\% & 91\%            & 32.0\%    \\
&                        & 4.1Gb  & 39.0\% & 95\%            & 39.9\%    \\
&                        & 5.5Gb    & 18.3\% & 98\%            & 48.7\%    \\

\midrule
\multirow{3}{*}{30B} 
& \multirow{3}{*}{13.1Gb}  & 5.7Gb   & 56.7\% & 93\%  & 37.0\%   \\
&                         & 6.7Gb   & 48.8\% & 95\%  & 42.5\%   \\
&                         & 9.5Gb  & 27.4\% & 98\%  & 47.5\%   \\

\midrule
\multirow{3}{*}{65B} 
 & \multirow{3}{*}{21.5Gb} & 9.4Gb & 56.3\% & 93\%            & 40.9\% \\
 &                      & 11.8Gb & 44.9\% & 95\%            & 44.2\% \\
 &                      & 13.8Gb & 36.0\% & 98\%            & 49.8\% \\

\bottomrule
\end{tabular}
}
\caption{Memory footprint reduction by {\ours}. We compared the memory consumption between models with full KV cache, and models compressed by {\ours} on fine-tuned Llama 1.}
\label{tab:memory_reduction}
\end{table*}

\subsection{Memory Footprint Reduction Analysis}
\label{subsec:memory}
We report the KV cache memory footprint reduction in Table \ref{tab:memory_reduction}.
For all the evaluated 7B-65B models, we evaluate the memory consumption with a fixed batch size of 16, sequence length of 512, and model weights in fp16 format.
We observe that {\ours} substantially reduces the KV cache memory footprint across all model sizes, with more significant reductions for larger models.
Taking a win rate over 45\% as little-to-no quality regression, {\ours} can achieve $\sim$40\% memory reduction in Llama 1-65B, $\sim$30\% in Llama 1-30B, $\sim$20\% in Llama 1-13B and Llama 1-7B. 


\subsection{End-to-end Latency Improvement}
\label{sec:latency}
\begin{table}[!htp]\centering
\caption{End-to-end latency comparison on Llama 1-7B.}
\label{tab:latency}
\resizebox{\columnwidth}{!}{
\begin{tabular}{l|cccc|ccc|cc|c}
\toprule
Batch size &\multicolumn{4}{c|}{1} &\multicolumn{3}{c|}{2} &\multicolumn{2}{c|}{8} &\multicolumn{1}{c}{16} \\
\cmidrule{1-11}
[prompt len, gen len] & \small[32,512] &\small[32,2048] &\small[32,8192] &\small[32,16384] &\small[512,32] &\small[512,512] &\small[4096,4096] &\small[512,512] &\small[4096,4096] &\small[512,512] \\
\midrule
HF &13.35 &57.37 &299 &799.14 &1.12 &19.16 &167.64 &23.44 &OOM &OOM \\
DS &11.58 &47.12 &201.23 &435.74 &0.79 &10.45 &91.04 &12.93 &127.94 &OOM \\
FastGen &11.21 &44.6 &179.43 &359.83 &0.73 &9.71 &76.93 &10.57 &82.16 &OOM \\
Speed-up(\%) over HF &16.03\% &22.30\% &40.00\% &55.00\% &34.80\% &49.30\% &54.10\% &54.90\% &- &OOM \\
Speed-up(\%) over DS &3.20\% &5.35\% &10.83\% &17.42\% &7.59\% &7.08\% &15.50\% &18.25\% &35.78\% &OOM \\
\bottomrule
\end{tabular}
}

\end{table}
To analyze the end-to-end speedup of FastGen, we present the end-to-end latency improvement over full-cache setting and a strong model acceleration baseline in Table \ref{tab:latency}.
In the experiment, we record the total duration in seconds, measured from the start of prompt encoding, until the end of generation as the end-to-end latency.
For the full-cache baseline, we adopt the widely used Hugging Face Accelerate (HF) \citep{accelerate}, denoted as HF in Table \ref{tab:latency}. 
For FastGen, we implemented a customized kernel to handle the KV cache pruning operation. 
Specifically, we adapt the kernel from Deepspeed (DS) \citep{Aminabadi2022DeepSpeedIE} by adding the KV cache sparsity operation. 
We include the Deepspeed performance for fair comparison, denoted as DS in Table \ref{tab:latency}. 
All methods are tested on the same Nvidia V100 GPUs.

As shown in Table \ref{tab:latency}, we can observe that FastGen achieves significant end-to-end speed-up across all the generation settings. For the least significant case, FastGen can have a decent $16.04\%$ latency improvement over the HF baseline on a short generation length of 512. In the best cases, we can achieve up to $55.0\%$ latency reduction over HF with FastGen at a generation length of 16k. 
We can also observe that the relative speedup is greater with longer generation length. For example, given batch size = 1, FastGen’s relative speed-up rises from $16.04\%$ to $55.0\%$, as the generation length grows from 512 to 16k.
When comparing FastGen to DeepSpeed, we can still observe significant speed-up that gets bigger with batch size and generation length. 
Considering DeepSpeed is a full-stack optimized inference system, where not only attention computation is optimized, there is still much room to further improve FastGen by polishing the sparsity kernel.
We leave this unique research and engineering challenge to future works.
\subsection{Profiling Cost}
\label{sec:profiling_cost}
\begin{table}[!htp]\centering
\caption{Profiling time of Llama 1-65B. The Overall Generation Duration is measured from the start of decoding to the end of the generation length. The Profiling Duration is measured from the start of the decoding until Fastgen finishes the policy search.}\label{tab: overhead}
\scriptsize
\resizebox{0.9\columnwidth}{!}{
\begin{tabular}{lrrrrr}
\toprule
\tabincell{r}{Generation \\Length} &\tabincell{r}{Overall Generation\\ Duration (s) }&\tabincell{r}{Profiling\\ Duration (s)} &\tabincell{r}{Decoding Time \\ Per Token (s)} &Profiling/Overall (\%) \\\midrule
128 &30.98 &0.11 &0.10 &0.35\% \\
256 &50.1 &0.11 &0.10 &0.21\% \\
512 &94.98 &0.11 &0.10 &0.12\% \\
1024 &157.43 &0.11 &0.10 &0.07\% \\
\bottomrule
\end{tabular}
}
\end{table}
To better understand the overhead of the profiling step, we compare the profiling time with the total generation time across different generation lengths. We present the result in Table \ref{tab: overhead}.

We can observe that the profiling time only accounts for a very small percentage of the total generation duration, up to $0.35\%$ in our tested cases. Also, the overhead decreases as the generation length increases, dropping to $0.07\%$ when the generation length comes to 1024.

In terms of extra memory usage, it’s mainly introduced by one of the compression strategies, $\mC_{\small \mbox{frequent}}$, which needs to store an extra cumulative sum of attention scores for each attention head. To provide a detailed analysis, for each layer, the dimension of the KV cache is $(\texttt{batch\_size}, \texttt{num\_of\_head}, \texttt{sequence\_len}, \texttt{hidden\_dimension})$, while the dimension of extra memory for the cumulative attention scores is $(\texttt{batch\_size}, \texttt{num\_of\_head}, \texttt{sequence\_len})$. Considering \texttt{hidden\_dimension} = 128 for all model sizes, the memory overhead is $1/128$=$0.78\%$ compared to storing KV cache only, which is a negligible cost.


\section{Conclusion}
We have presented \ours, a novel method that significantly improves the inference efficiency of LLMs, with no visible quality loss, using lightweight model profiling and adaptive key-value caching. 
Areas for future explorations include combining {\ours} with other model compression techniques, such as quantization and distillation, and 
other efficient attention architectures, such as grouped-query attention.
\clearpage

\section*{Acknowledgments}
Research was supported in part by US DARPA KAIROS Program No. FA8750-19-2-1004 and INCAS Program No. HR001121C0165, National Science Foundation IIS-19-56151, and the Molecule Maker Lab Institute: An AI Research Institutes program supported by NSF under Award No. 2019897, and the Institute for Geospatial Understanding through an Integrative Discovery Environment (I-GUIDE) by NSF under Award No. 2118329. 
Any opinions, findings, and conclusions or recommendations expressed herein are those of the authors and do not necessarily represent the views, either expressed or implied, of DARPA or the U.S. Government.


\bibliography{iclr2024_conference}

\begin{thebibliography}{38}
\providecommand{\natexlab}[1]{#1}
\providecommand{\url}[1]{\texttt{#1}}
\expandafter\ifx\csname urlstyle\endcsname\relax
  \providecommand{\doi}[1]{doi: #1}\else
  \providecommand{\doi}{doi: \begingroup \urlstyle{rm}\Url}\fi

\bibitem[Aminabadi et~al.(2022)Aminabadi, Rajbhandari, Zhang, Awan, Li, Li,
  Zheng, Rasley, Smith, Ruwase, and He]{Aminabadi2022DeepSpeedIE}
Reza~Yazdani Aminabadi, Samyam Rajbhandari, Minjia Zhang, Ammar~Ahmad Awan,
  Cheng Li, Du~Li, Elton Zheng, Jeff Rasley, Shaden Smith, Olatunji Ruwase, and
  Yuxiong He.
\newblock Deepspeed- inference: Enabling efficient inference of transformer
  models at unprecedented scale.
\newblock \emph{SC22: International Conference for High Performance Computing,
  Networking, Storage and Analysis}, pp.\  1--15, 2022.

\bibitem[Bach et~al.(2015)Bach, Binder, Montavon, Klauschen, M{\"u}ller, and
  Samek]{lrf}
Sebastian Bach, Alexander Binder, Gr{\'e}goire Montavon, Frederick Klauschen,
  Klaus-Robert M{\"u}ller, and Wojciech Samek.
\newblock On pixel-wise explanations for non-linear classifier decisions by
  layer-wise relevance propagation.
\newblock \emph{PLoS ONE}, 10, 2015.
\newblock URL \url{https://api.semanticscholar.org/CorpusID:9327892}.

\bibitem[Brown et~al.(2020)Brown, Mann, Ryder, Subbiah, Kaplan, Dhariwal,
  Neelakantan, Shyam, Sastry, Askell, Agarwal, Herbert{-}Voss, Krueger,
  Henighan, Child, Ramesh, Ziegler, Wu, Winter, Hesse, Chen, Sigler, Litwin,
  Gray, Chess, Clark, Berner, McCandlish, Radford, Sutskever, and
  Amodei]{gpt-3}
Tom~B. Brown, Benjamin Mann, Nick Ryder, Melanie Subbiah, Jared Kaplan,
  Prafulla Dhariwal, Arvind Neelakantan, Pranav Shyam, Girish Sastry, Amanda
  Askell, Sandhini Agarwal, Ariel Herbert{-}Voss, Gretchen Krueger, Tom
  Henighan, Rewon Child, Aditya Ramesh, Daniel~M. Ziegler, Jeffrey Wu, Clemens
  Winter, Christopher Hesse, Mark Chen, Eric Sigler, Mateusz Litwin, Scott
  Gray, Benjamin Chess, Jack Clark, Christopher Berner, Sam McCandlish, Alec
  Radford, Ilya Sutskever, and Dario Amodei.
\newblock Language models are few-shot learners.
\newblock In Hugo Larochelle, Marc'Aurelio Ranzato, Raia Hadsell,
  Maria{-}Florina Balcan, and Hsuan{-}Tien Lin (eds.), \emph{Advances in Neural
  Information Processing Systems 33: Annual Conference on Neural Information
  Processing Systems 2020, NeurIPS 2020, December 6-12, 2020, virtual}, 2020.

\bibitem[Campos et~al.(2017)Campos, Jou, i~Nieto, Torres, and
  Chang]{Campos2017SkipRL}
V{\'i}ctor Campos, Brendan Jou, Xavier~Gir{\'o} i~Nieto, Jordi Torres, and
  Shih-Fu Chang.
\newblock Skip rnn: Learning to skip state updates in recurrent neural
  networks.
\newblock \emph{ArXiv}, abs/1708.06834, 2017.
\newblock URL \url{https://api.semanticscholar.org/CorpusID:1859294}.

\bibitem[Chen et~al.(2021)Chen, Tworek, Jun, Yuan, Pinto, Kaplan, Edwards,
  Burda, Joseph, Brockman, et~al.]{human_eval}
Mark Chen, Jerry Tworek, Heewoo Jun, Qiming Yuan, Henrique Ponde de~Oliveira
  Pinto, Jared Kaplan, Harri Edwards, Yuri Burda, Nicholas Joseph, Greg
  Brockman, et~al.
\newblock Evaluating large language models trained on code.
\newblock \emph{arXiv preprint arXiv:2107.03374}, 2021.

\bibitem[Child et~al.(2019)Child, Gray, Radford, and
  Sutskever]{sparsetransformer}
Rewon Child, Scott Gray, Alec Radford, and Ilya Sutskever.
\newblock Generating long sequences with sparse transformers.
\newblock \emph{CoRR}, abs/1904.10509, 2019.
\newblock URL \url{http://arxiv.org/abs/1904.10509}.

\bibitem[Clark et~al.(2019)Clark, Khandelwal, Levy, and Manning]{bert-look}
Kevin Clark, Urvashi Khandelwal, Omer Levy, and Christopher~D. Manning.
\newblock What does {BERT} look at? an analysis of {BERT}{'}s attention.
\newblock In \emph{Proceedings of the 2019 ACL Workshop BlackboxNLP: Analyzing
  and Interpreting Neural Networks for NLP}, pp.\  276--286, Florence, Italy,
  August 2019. Association for Computational Linguistics.
\newblock \doi{10.18653/v1/W19-4828}.
\newblock URL \url{https://aclanthology.org/W19-4828}.

\bibitem[Cobbe et~al.(2021)Cobbe, Kosaraju, Bavarian, Chen, Jun, Kaiser,
  Plappert, Tworek, Hilton, Nakano, et~al.]{gsm8k}
Karl Cobbe, Vineet Kosaraju, Mohammad Bavarian, Mark Chen, Heewoo Jun, Lukasz
  Kaiser, Matthias Plappert, Jerry Tworek, Jacob Hilton, Reiichiro Nakano,
  et~al.
\newblock Training verifiers to solve math word problems.
\newblock \emph{arXiv preprint arXiv:2110.14168}, 2021.

\bibitem[Dai et~al.(2020)Dai, Lai, Yang, and Le]{funnel}
Zihang Dai, Guokun Lai, Yiming Yang, and Quoc Le.
\newblock Funnel-transformer: Filtering out sequential redundancy for efficient
  language processing.
\newblock In Hugo Larochelle, Marc'Aurelio Ranzato, Raia Hadsell,
  Maria{-}Florina Balcan, and Hsuan{-}Tien Lin (eds.), \emph{Advances in Neural
  Information Processing Systems 33: Annual Conference on Neural Information
  Processing Systems 2020, NeurIPS 2020, December 6-12, 2020, virtual}, 2020.
\newblock URL
  \url{https://proceedings.neurips.cc/paper/2020/hash/2cd2915e69546904e4e5d4a2ac9e1652-Abstract.html}.

\bibitem[Devlin et~al.(2019)Devlin, Chang, Lee, and Toutanova]{bert}
Jacob Devlin, Ming{-}Wei Chang, Kenton Lee, and Kristina Toutanova.
\newblock {BERT:} pre-training of deep bidirectional transformers for language
  understanding.
\newblock In Jill Burstein, Christy Doran, and Thamar Solorio (eds.),
  \emph{Proceedings of the 2019 Conference of the North American Chapter of the
  Association for Computational Linguistics: Human Language Technologies,
  {NAACL-HLT} 2019, Minneapolis, MN, USA, June 2-7, 2019, Volume 1 (Long and
  Short Papers)}, pp.\  4171--4186. Association for Computational Linguistics,
  2019.
\newblock \doi{10.18653/v1/n19-1423}.
\newblock URL \url{https://doi.org/10.18653/v1/n19-1423}.

\bibitem[Goyal et~al.(2020)Goyal, Choudhury, Raje, Chakaravarthy, Sabharwal,
  and Verma]{powerbert}
Saurabh Goyal, Anamitra~R. Choudhury, Saurabh Raje, Venkatesan~T.
  Chakaravarthy, Yogish Sabharwal, and Ashish Verma.
\newblock Power-bert: Accelerating bert inference via progressive word-vector
  elimination.
\newblock In \emph{International Conference on Machine Learning}, 2020.
\newblock URL \url{https://api.semanticscholar.org/CorpusID:219792793}.

\bibitem[Guan et~al.(2022)Guan, Li, Leng, Lin, and Guo]{transkimmer}
Yue Guan, Zhengyi Li, Jingwen Leng, Zhouhan Lin, and Minyi Guo.
\newblock Transkimmer: Transformer learns to layer-wise skim.
\newblock In Smaranda Muresan, Preslav Nakov, and Aline Villavicencio (eds.),
  \emph{Proceedings of the 60th Annual Meeting of the Association for
  Computational Linguistics (Volume 1: Long Papers), {ACL} 2022, Dublin,
  Ireland, May 22-27, 2022}, pp.\  7275--7286. Association for Computational
  Linguistics, 2022.
\newblock \doi{10.18653/v1/2022.acl-long.502}.
\newblock URL \url{https://doi.org/10.18653/v1/2022.acl-long.502}.

\bibitem[Gugger et~al.(2022)Gugger, Debut, Wolf, Schmid, Mueller, Mangrulkar,
  Sun, and Bossan]{accelerate}
Sylvain Gugger, Lysandre Debut, Thomas Wolf, Philipp Schmid, Zachary Mueller,
  Sourab Mangrulkar, Marc Sun, and Benjamin Bossan.
\newblock Accelerate: Training and inference at scale made simple, efficient
  and adaptable.
\newblock \url{https://github.com/huggingface/accelerate}, 2022.

\bibitem[Hansen et~al.(2019)Hansen, Hansen, Alstrup, Simonsen, and
  Lioma]{Hansen2019NeuralSR}
Christian Hansen, Casper Hansen, Stephen Alstrup, Jakob~Grue Simonsen, and
  Christina Lioma.
\newblock Neural speed reading with structural-jump-lstm.
\newblock \emph{ArXiv}, abs/1904.00761, 2019.
\newblock URL \url{https://api.semanticscholar.org/CorpusID:90258012}.

\bibitem[Holtzman et~al.(2019)Holtzman, Buys, Du, Forbes, and
  Choi]{holtzman2019curious}
Ari Holtzman, Jan Buys, Li~Du, Maxwell Forbes, and Yejin Choi.
\newblock The curious case of neural text degeneration.
\newblock \emph{arXiv preprint arXiv:1904.09751}, 2019.

\bibitem[Huang et~al.(2022)Huang, Khetan, Bidart, and Karnin]{pyramid}
Xin Huang, Ashish Khetan, Rene Bidart, and Zohar Karnin.
\newblock Pyramid-bert: Reducing complexity via successive core-set based token
  selection.
\newblock In Smaranda Muresan, Preslav Nakov, and Aline Villavicencio (eds.),
  \emph{Proceedings of the 60th Annual Meeting of the Association for
  Computational Linguistics (Volume 1: Long Papers), {ACL} 2022, Dublin,
  Ireland, May 22-27, 2022}, pp.\  8798--8817. Association for Computational
  Linguistics, 2022.
\newblock \doi{10.18653/v1/2022.acl-long.602}.
\newblock URL \url{https://doi.org/10.18653/v1/2022.acl-long.602}.

\bibitem[Kembhavi et~al.(2017)Kembhavi, Seo, Schwenk, Choi, Farhadi, and
  Hajishirzi]{tqa}
Aniruddha Kembhavi, Minjoon Seo, Dustin Schwenk, Jonghyun Choi, Ali Farhadi,
  and Hannaneh Hajishirzi.
\newblock Are you smarter than a sixth grader? textbook question answering for
  multimodal machine comprehension.
\newblock \emph{2017 IEEE Conference on Computer Vision and Pattern Recognition
  (CVPR)}, pp.\  5376--5384, 2017.
\newblock URL \url{https://api.semanticscholar.org/CorpusID:1310550}.

\bibitem[Kim et~al.(2022)Kim, Shen, Thorsley, Gholami, Kwon, Hassoun, and
  Keutzer]{tokenprune}
Sehoon Kim, Sheng Shen, David Thorsley, Amir Gholami, Woosuk Kwon, Joseph
  Hassoun, and Kurt Keutzer.
\newblock Learned token pruning for transformers.
\newblock In Aidong Zhang and Huzefa Rangwala (eds.), \emph{{KDD} '22: The 28th
  {ACM} {SIGKDD} Conference on Knowledge Discovery and Data Mining, Washington,
  DC, USA, August 14 - 18, 2022}, pp.\  784--794. {ACM}, 2022.
\newblock \doi{10.1145/3534678.3539260}.
\newblock URL \url{https://doi.org/10.1145/3534678.3539260}.

\bibitem[K{\"{o}}pf et~al.(2023)K{\"{o}}pf, Kilcher, von R{\"{u}}tte,
  Anagnostidis, Tam, Stevens, Barhoum, Duc, Stanley, Nagyfi, ES, Suri,
  Glushkov, Dantuluri, Maguire, Schuhmann, Nguyen, and Mattick]{openass}
Andreas K{\"{o}}pf, Yannic Kilcher, Dimitri von R{\"{u}}tte, Sotiris
  Anagnostidis, Zhi{-}Rui Tam, Keith Stevens, Abdullah Barhoum, Nguyen~Minh
  Duc, Oliver Stanley, Rich{\'{a}}rd Nagyfi, Shahul ES, Sameer Suri, David
  Glushkov, Arnav Dantuluri, Andrew Maguire, Christoph Schuhmann, Huu Nguyen,
  and Alexander Mattick.
\newblock Openassistant conversations - democratizing large language model
  alignment.
\newblock \emph{CoRR}, abs/2304.07327, 2023.
\newblock \doi{10.48550/arXiv.2304.07327}.
\newblock URL \url{https://doi.org/10.48550/arXiv.2304.07327}.

\bibitem[Kovaleva et~al.(2019)Kovaleva, Romanov, Rogers, and Rumshisky]{dark}
Olga Kovaleva, Alexey Romanov, Anna Rogers, and Anna Rumshisky.
\newblock Revealing the dark secrets of {BERT}.
\newblock In Kentaro Inui, Jing Jiang, Vincent Ng, and Xiaojun Wan (eds.),
  \emph{Proceedings of the 2019 Conference on Empirical Methods in Natural
  Language Processing and the 9th International Joint Conference on Natural
  Language Processing, {EMNLP-IJCNLP} 2019, Hong Kong, China, November 3-7,
  2019}, pp.\  4364--4373. Association for Computational Linguistics, 2019.
\newblock \doi{10.18653/v1/D19-1445}.
\newblock URL \url{https://doi.org/10.18653/v1/D19-1445}.

\bibitem[Kwiatkowski et~al.(2019)Kwiatkowski, Palomaki, Redfield, Collins,
  Parikh, Alberti, Epstein, Polosukhin, Kelcey, Devlin, Lee, Toutanova, Jones,
  Chang, Dai, Uszkoreit, Le, and Petrov]{nq}
Tom Kwiatkowski, Jennimaria Palomaki, Olivia Redfield, Michael Collins, Ankur
  Parikh, Chris Alberti, Danielle Epstein, Illia Polosukhin, Matthew Kelcey,
  Jacob Devlin, Kenton Lee, Kristina~N. Toutanova, Llion Jones, Ming-Wei Chang,
  Andrew Dai, Jakob Uszkoreit, Quoc Le, and Slav Petrov.
\newblock Natural questions: a benchmark for question answering research.
\newblock \emph{Transactions of the Association of Computational Linguistics},
  2019.

\bibitem[Li et~al.(2023)Li, Zhang, Dubois, Taori, Gulrajani, Guestrin, Liang,
  and Hashimoto]{alpaca_eval}
Xuechen Li, Tianyi Zhang, Yann Dubois, Rohan Taori, Ishaan Gulrajani, Carlos
  Guestrin, Percy Liang, and Tatsunori~B. Hashimoto.
\newblock Alpacaeval: An automatic evaluator of instruction-following models.
\newblock \url{https://github.com/tatsu-lab/alpaca_eval}, 2023.

\bibitem[Liu et~al.(2023{\natexlab{a}})Liu, Desai, Liao, Wang, Xie, Xu,
  Kyrillidis, and Shrivastava]{scissorhands}
Zichang Liu, Aditya Desai, Fangshuo Liao, Weitao Wang, Victor Xie, Zhaozhuo Xu,
  Anastasios Kyrillidis, and Anshumali Shrivastava.
\newblock Scissorhands: Exploiting the persistence of importance hypothesis for
  {LLM} {KV} cache compression at test time.
\newblock \emph{CoRR}, abs/2305.17118, 2023{\natexlab{a}}.
\newblock \doi{10.48550/arXiv.2305.17118}.
\newblock URL \url{https://doi.org/10.48550/arXiv.2305.17118}.

\bibitem[Liu et~al.(2023{\natexlab{b}})Liu, Wang, Dao, Zhou, Yuan, Song,
  Shrivastava, Zhang, Tian, R{\'{e}}, and Chen]{deja}
Zichang Liu, Jue Wang, Tri Dao, Tianyi Zhou, Binhang Yuan, Zhao Song, Anshumali
  Shrivastava, Ce~Zhang, Yuandong Tian, Christopher R{\'{e}}, and Beidi Chen.
\newblock Deja vu: Contextual sparsity for efficient llms at inference time.
\newblock In Andreas Krause, Emma Brunskill, Kyunghyun Cho, Barbara Engelhardt,
  Sivan Sabato, and Jonathan Scarlett (eds.), \emph{International Conference on
  Machine Learning, {ICML} 2023, 23-29 July 2023, Honolulu, Hawaii, {USA}},
  volume 202 of \emph{Proceedings of Machine Learning Research}, pp.\
  22137--22176. {PMLR}, 2023{\natexlab{b}}.
\newblock URL \url{https://proceedings.mlr.press/v202/liu23am.html}.

\bibitem[Michel et~al.(2019)Michel, Levy, and Neubig]{sixteen}
Paul Michel, Omer Levy, and Graham Neubig.
\newblock Are sixteen heads really better than one?
\newblock In H.~Wallach, H.~Larochelle, A.~Beygelzimer, F.~d\textquotesingle
  Alch\'{e}-Buc, E.~Fox, and R.~Garnett (eds.), \emph{Advances in Neural
  Information Processing Systems}, volume~32. Curran Associates, Inc., 2019.
\newblock URL
  \url{https://proceedings.neurips.cc/paper_files/paper/2019/file/2c601ad9d2ff9bc8b282670cdd54f69f-Paper.pdf}.

\bibitem[Mu et~al.(2023)Mu, Li, and Goodman]{gist}
Jesse Mu, Xiang~Lisa Li, and Noah~D. Goodman.
\newblock Learning to compress prompts with gist tokens.
\newblock \emph{CoRR}, abs/2304.08467, 2023.
\newblock \doi{10.48550/arXiv.2304.08467}.
\newblock URL \url{https://doi.org/10.48550/arXiv.2304.08467}.

\bibitem[OpenAI(2023)]{openai2023gpt4}
OpenAI.
\newblock Gpt-4 technical report, 2023.

\bibitem[Seo et~al.(2017)Seo, Min, Farhadi, and Hajishirzi]{Seo2017NeuralSR}
Minjoon Seo, Sewon Min, Ali Farhadi, and Hannaneh Hajishirzi.
\newblock Neural speed reading via skim-rnn.
\newblock \emph{ArXiv}, abs/1711.02085, 2017.
\newblock URL \url{https://api.semanticscholar.org/CorpusID:3140413}.

\bibitem[Shazeer et~al.(2017)Shazeer, Mirhoseini, Maziarz, Davis, Le, Hinton,
  and Dean]{540b}
Noam~M. Shazeer, Azalia Mirhoseini, Krzysztof Maziarz, Andy Davis, Quoc~V. Le,
  Geoffrey~E. Hinton, and Jeff Dean.
\newblock Outrageously large neural networks: The sparsely-gated
  mixture-of-experts layer.
\newblock \emph{ArXiv}, abs/1701.06538, 2017.
\newblock URL \url{https://api.semanticscholar.org/CorpusID:12462234}.

\bibitem[Sheng et~al.(2023)Sheng, Zheng, Yuan, Li, Ryabinin, Fu, Xie, Chen,
  Barrett, Gonzalez, Liang, R{\'e}, Stoica, and
  Zhang]{Sheng2023HighthroughputGI}
Ying Sheng, Lianmin Zheng, Binhang Yuan, Zhuohan Li, Max Ryabinin, Daniel~Y.
  Fu, Zhiqiang Xie, Beidi Chen, Clark~W. Barrett, Joseph Gonzalez, Percy Liang,
  Christopher R{\'e}, Ioan~Cristian Stoica, and Ce~Zhang.
\newblock High-throughput generative inference of large language models with a
  single gpu.
\newblock In \emph{International Conference on Machine Learning}, 2023.
\newblock URL \url{https://api.semanticscholar.org/CorpusID:257495837}.

\bibitem[Sun et~al.(2022)Sun, Liu, Zhu, Geng, Wu, He, Ni, Xie, Huang, and
  Qiu]{hashexit}
Tianxiang Sun, Xiangyang Liu, Wei Zhu, Zhichao Geng, Lingling Wu, Yilong He,
  Yuan Ni, Guotong Xie, Xuanjing Huang, and Xipeng Qiu.
\newblock A simple hash-based early exiting approach for language understanding
  and generation.
\newblock In Smaranda Muresan, Preslav Nakov, and Aline Villavicencio (eds.),
  \emph{Findings of the Association for Computational Linguistics: {ACL} 2022,
  Dublin, Ireland, May 22-27, 2022}, pp.\  2409--2421. Association for
  Computational Linguistics, 2022.
\newblock \doi{10.18653/v1/2022.findings-acl.189}.
\newblock URL \url{https://doi.org/10.18653/v1/2022.findings-acl.189}.

\bibitem[Touvron et~al.(2023{\natexlab{a}})Touvron, Lavril, Izacard, Martinet,
  Lachaux, Lacroix, Rozi{\`e}re, Goyal, Hambro, Azhar,
  et~al.]{touvron2023llama1}
Hugo Touvron, Thibaut Lavril, Gautier Izacard, Xavier Martinet, Marie-Anne
  Lachaux, Timoth{\'e}e Lacroix, Baptiste Rozi{\`e}re, Naman Goyal, Eric
  Hambro, Faisal Azhar, et~al.
\newblock Llama: Open and efficient foundation language models.
\newblock \emph{arXiv preprint arXiv:2302.13971}, 2023{\natexlab{a}}.

\bibitem[Touvron et~al.(2023{\natexlab{b}})Touvron, Martin, Stone, Albert,
  Almahairi, Babaei, Bashlykov, Batra, Bhargava, Bhosale, Bikel, Blecher,
  Ferrer, Chen, Cucurull, Esiobu, Fernandes, Fu, Fu, Fuller, Gao, Goswami,
  Goyal, Hartshorn, Hosseini, Hou, Inan, Kardas, Kerkez, Khabsa, Kloumann,
  Korenev, Koura, Lachaux, Lavril, Lee, Liskovich, Lu, Mao, Martinet, Mihaylov,
  Mishra, Molybog, Nie, Poulton, Reizenstein, Rungta, Saladi, Schelten, Silva,
  Smith, Subramanian, Tan, Tang, Taylor, Williams, Kuan, Xu, Yan, Zarov, Zhang,
  Fan, Kambadur, Narang, Rodriguez, Stojnic, Edunov, and
  Scialom]{touvron2023llama}
Hugo Touvron, Louis Martin, Kevin Stone, Peter Albert, Amjad Almahairi, Yasmine
  Babaei, Nikolay Bashlykov, Soumya Batra, Prajjwal Bhargava, Shruti Bhosale,
  Dan Bikel, Lukas Blecher, Cristian~Canton Ferrer, Moya Chen, Guillem
  Cucurull, David Esiobu, Jude Fernandes, Jeremy Fu, Wenyin Fu, Brian Fuller,
  Cynthia Gao, Vedanuj Goswami, Naman Goyal, Anthony Hartshorn, Saghar
  Hosseini, Rui Hou, Hakan Inan, Marcin Kardas, Viktor Kerkez, Madian Khabsa,
  Isabel Kloumann, Artem Korenev, Punit~Singh Koura, Marie-Anne Lachaux,
  Thibaut Lavril, Jenya Lee, Diana Liskovich, Yinghai Lu, Yuning Mao, Xavier
  Martinet, Todor Mihaylov, Pushkar Mishra, Igor Molybog, Yixin Nie, Andrew
  Poulton, Jeremy Reizenstein, Rashi Rungta, Kalyan Saladi, Alan Schelten, Ruan
  Silva, Eric~Michael Smith, Ranjan Subramanian, Xiaoqing~Ellen Tan, Binh Tang,
  Ross Taylor, Adina Williams, Jian~Xiang Kuan, Puxin Xu, Zheng Yan, Iliyan
  Zarov, Yuchen Zhang, Angela Fan, Melanie Kambadur, Sharan Narang, Aurelien
  Rodriguez, Robert Stojnic, Sergey Edunov, and Thomas Scialom.
\newblock Llama 2: Open foundation and fine-tuned chat models,
  2023{\natexlab{b}}.

\bibitem[Voita et~al.(2019)Voita, Talbot, Moiseev, Sennrich, and
  Titov]{lift-prune}
Elena Voita, David Talbot, Fedor Moiseev, Rico Sennrich, and Ivan Titov.
\newblock Analyzing multi-head self-attention: Specialized heads do the heavy
  lifting, the rest can be pruned, July 2019.
\newblock URL \url{https://aclanthology.org/P19-1580}.

\bibitem[Wang et~al.(2020)Wang, Wohlwend, and Lei]{structureprune}
Ziheng Wang, Jeremy Wohlwend, and Tao Lei.
\newblock Structured pruning of large language models.
\newblock In Bonnie Webber, Trevor Cohn, Yulan He, and Yang Liu (eds.),
  \emph{Proceedings of the 2020 Conference on Empirical Methods in Natural
  Language Processing, {EMNLP} 2020, Online, November 16-20, 2020}, pp.\
  6151--6162. Association for Computational Linguistics, 2020.
\newblock \doi{10.18653/v1/2020.emnlp-main.496}.
\newblock URL \url{https://doi.org/10.18653/v1/2020.emnlp-main.496}.

\bibitem[Zhang et~al.(2023)Zhang, Sheng, Zhou, Chen, Zheng, Cai, Song, Tian,
  R{\'{e}}, Barrett, Wang, and Chen]{h2o}
Zhenyu Zhang, Ying Sheng, Tianyi Zhou, Tianlong Chen, Lianmin Zheng, Ruisi Cai,
  Zhao Song, Yuandong Tian, Christopher R{\'{e}}, Clark~W. Barrett, Zhangyang
  Wang, and Beidi Chen.
\newblock H\({}_{\mbox{2}}\)o: Heavy-hitter oracle for efficient generative
  inference of large language models.
\newblock \emph{CoRR}, abs/2306.14048, 2023.
\newblock \doi{10.48550/arXiv.2306.14048}.
\newblock URL \url{https://doi.org/10.48550/arXiv.2306.14048}.

\bibitem[Zhou et~al.(2023)Zhou, Liu, Xu, Iyer, Sun, Mao, Ma, Efrat, Yu, Yu,
  Zhang, Ghosh, Lewis, Zettlemoyer, and Levy]{lima}
Chunting Zhou, Pengfei Liu, Puxin Xu, Srini Iyer, Jiao Sun, Yuning Mao, Xuezhe
  Ma, Avia Efrat, Ping Yu, Lili Yu, Susan Zhang, Gargi Ghosh, Mike Lewis, Luke
  Zettlemoyer, and Omer Levy.
\newblock {LIMA:} less is more for alignment.
\newblock \emph{CoRR}, abs/2305.11206, 2023.
\newblock \doi{10.48550/arXiv.2305.11206}.
\newblock URL \url{https://doi.org/10.48550/arXiv.2305.11206}.

\bibitem[Zhou et~al.(2020)Zhou, Xu, Ge, McAuley, Xu, and Wei]{patience}
Wangchunshu Zhou, Canwen Xu, Tao Ge, Julian~J. McAuley, Ke~Xu, and Furu Wei.
\newblock {BERT} loses patience: Fast and robust inference with early exit.
\newblock In Hugo Larochelle, Marc'Aurelio Ranzato, Raia Hadsell,
  Maria{-}Florina Balcan, and Hsuan{-}Tien Lin (eds.), \emph{Advances in Neural
  Information Processing Systems 33: Annual Conference on Neural Information
  Processing Systems 2020, NeurIPS 2020, December 6-12, 2020, virtual}, 2020.
\newblock URL
  \url{https://proceedings.neurips.cc/paper/2020/hash/d4dd111a4fd973394238aca5c05bebe3-Abstract.html}.

\end{thebibliography}
\bibliographystyle{iclr2024_conference}
\appendix
\section{Appendix}
\subsection{Ablations}
\label{subsec:ablation}
For all the ablations, we use a fixed targeted recovery ratio T $=0.98$.

\paragraph{How one policy affect all the other policies?}
We study the complementary effects of each policy on the combination of all other policies in our framework.
We examine changes in pruned KV cache and win rate while fixing the targeted recovery ratio $T$. 
We take the full policy set as our control set $\gC$.
For each ablation, we remove one of the policies from all policy combination in $\gC$. 
We summarized the results in Table \ref{tab:cache_type_abl}, which suggests the $\mC_{\small \mbox{frequent}}$, and the $\mC_{\small \mbox{special}}$ are the most important policies. 
Removing them will incur a $3.67\%$ and a $2.11\%$ win rate drop respectively.
We can also observe from the pruned cache ratio that $\mC_{\small \mbox{frequent}}$ and $\mC_{\small \mbox{local}}$ reduce more KV caches than the others. 
However, their standalone non-adaptive deployment yields suboptimal performance, as depicted in Figure \ref{fig:exp_main}, 
further verifying the importance of adapting different compression policies.

\begin{table}[t]
\small
\centering
\resizebox{0.9\columnwidth}{!}{
    \begin{tabular}{l cc}
       \toprule

    \textbf{Feasible Policy Set} & \textbf{Pruned KV Ratio} & \textbf{Win Rate}\\
       \midrule
        $\gC$ & 36.04\% & 49.75\% \\
       \midrule

        $\{\mC_{\small \mbox{punct.}}, \mC_{\small \mbox{punct.+frequent}}, \mC_{\small \mbox{punct.+frequent+local}}, \mC_{\small \mbox{full}} \}$ &31.16\% &47.64\% \\
       \midrule
       $\{\mC_{\small \mbox{special}}, \mC_{\small \mbox{special+frequent}}, \mC_{\small \mbox{special+frequent+local}}, \mC_{\small \mbox{full}} \}$ &34.23\% &49.56\% \\
       \midrule
       $\{\mC_{\small \mbox{special}}, \mC_{\small \mbox{special+punct.}}, \mC_{\small \mbox{special+punct.+frequent}}, \mC_{\small \mbox{full}} \}$ &30.18\% &49.06\% \\
       \midrule
       $\{\mC_{\small \mbox{special}}, \mC_{\small \mbox{special+punct.}}, \mC_{\small \mbox{special+punct.+local}}, \mC_{\small \mbox{full}} \}$ &21.26\% &46.08\% \\

       \bottomrule
    \end{tabular}
}
	\caption{Complementary effects of each policy. We display the win rate of each method over full cache setting. We evaluate the fine-tuned Llama 1-65B on AlpacaEval with the same parameters.}
        \label{tab:cache_type_abl}
\end{table}


\begin{table}[t]
    \small
    \centering
    \resizebox{0.75\columnwidth}{!}{
            \begin{tabular}{l|cc}
               \toprule
            \textbf{Cache Order} & \textbf{Pruned KV Ratio} & \textbf{Win Rate}\\
               \midrule
               $\mC_{\small \mbox{special}}\to\mC_{\small \mbox{punct.}}\to \mC_{\small \mbox{frequent}}\to \mC_{\small \mbox{local}}$ &36.04\% &49.75\% \\
               \midrule
               $\mC_{\small \mbox{special}}\to\mC_{\small \mbox{frequent}}\to \mC_{\small \mbox{local}}\to \mC_{\small \mbox{punct.}}$&36.40\% &47.64\% \\
               \bottomrule
            \end{tabular}
      }
        \caption{Policy order ablation on fine-tuned Llama 1-65B with AlpacaEval.}
            \label{tab:cache_order_abl}
    \end{table}
\begin{figure}[h!]
    \includegraphics[width=\textwidth]{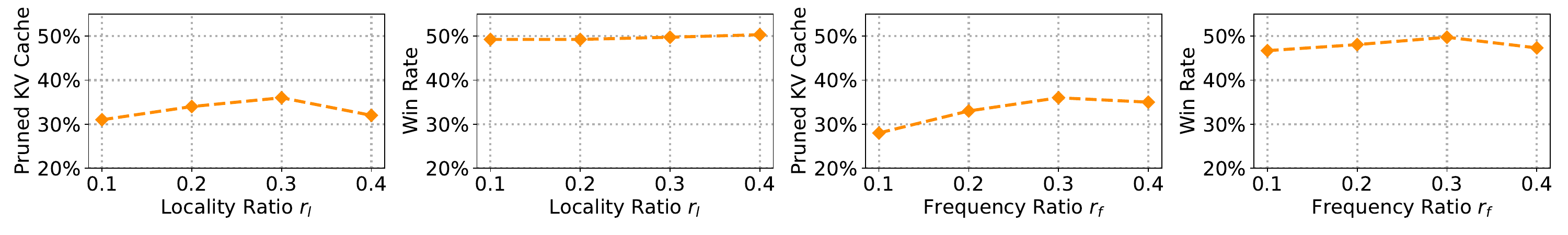}
    \caption{Hyper-parameter ablation on fine-tuned Llama 1-65B with AlpacaEval.}
    \label{fig:exp_hyperpar}
\end{figure}
\vspace{-0.3cm}

\paragraph{Which policy should we add first (and last)?}
As in Section~\ref{subsec:compression_policy}, we use a greed method to construct adaptive KV cache. 
Here, we examine how the order of introducing each policy affects the performance.
Similar to the previous study, we fix the targeted recovery ratio to 0.98, and keep allocating cache budget until the constructed cache hit the recovery ratio.
For simplicity, we make every examined order opt-in the $\mC_{\small \mbox{special}}$ first, as it's typically the most important tokens and of super-low memory cost, as suggested in Figure \ref{fig:obs}.
We summarize the results in Table \ref{tab:cache_order_abl}.
Our current order (as in Equation \ref{eqn:policyset}) achieves the highest win-rates and the highest pruned ratios. 
Meanwhile, using alternative orders leads to a different trade-off between KV cache compression and generation quality. 
For example, using $\mC_{\small \mbox{frequent}}\rightarrow \mC_{\small \mbox{local}}\rightarrow \mC_{\small \mbox{punct.}}$ leads to an improved KV cache compression ratio at the cost of generation quality.

\subsection{Sensitivity Study.}
We analyze the sensitivity of selecting different hyper-parameters for {\ours}, as illustrated in Figure \ref{fig:exp_hyperpar}. 
We observe that altering these hyper-parameters does not have a visible impact on the generation quality, as  
the model maintains a winrate over 45\% in all situations. 
Meanwhile, it leads to a relative large change on the compression ratio. 
For example, changing the ratio for the frequency policy from 0.3 to 0.1 leads to more KV cache. 
In our experiments, we set the ratio to 0.3 for both $r_l$ and $r_f$. 
\end{document}